\documentclass[runningheads]{llncs}

\usepackage{todonotes}
\usepackage{paralist}
\usepackage{siunitx}

\usepackage[utf8]{inputenc} %
\usepackage[T1]{fontenc}    %
\usepackage{url}            %
\usepackage{booktabs}       %
\usepackage{amsfonts}       %
\usepackage{nicefrac}       %
\usepackage{microtype}      %
\usepackage{xcolor}         %
\usepackage{comment}

\usepackage{booktabs} 
\usepackage{caption} 
\usepackage{graphicx} %
\usepackage{pgfplots} %
\usepackage{tikz} %
\usetikzlibrary{shapes,arrows}
\usepackage{pgfplots}
\usepackage{float}
\usepackage{bussproofs}
\usepackage{xspace}
\usepackage{amsmath}
\usepackage{mathtools}
\usepackage{doi}
\usepackage{multicol}
\usepackage{mathpartir}
\usepackage{verbatimbox}

\def\isabellehol{\textsf{Isabelle/HOL}\xspace}

\def\coq{\textsf{Coq}\xspace}

\def\vampire{\textsf{Vampire}\xspace}

\def\cvc{\textsf{CVC5}\xspace}
\def\zthree{\textsf{Z3}\xspace}
\def\sml{\textsf{SML}\xspace}

\def\sml{\textsf{Standard ML}\xspace}
\usepackage{color}

\usetikzlibrary{arrows,shapes,calc}
\usetikzlibrary{trees,positioning,fit}
\tikzstyle{arrow}=[draw,-to,thick]
\tikzstyle{embedding} = [draw, minimum width=8mm, minimum height=6mm]
\tikzstyle{nnop} = [draw, minimum width=8mm, minimum height=8mm, rounded 
corners]
\tikzstyle{block} =
[rectangle, draw,
minimum width=9mm, text centered, rounded corners, minimum height=2.0em]
\tikzstyle{smallblock} =
[rectangle, draw,
minimum width=9mm, text centered, minimum height=2.0em]

\tikzstyle{line}=[draw]
\tikzstyle{cloud} =
[draw, text centered, ellipse, minimum height=2.0em, minimum width=9em]

\usepackage[]{listings}

\usepackage{xcolor}

\lstdefinelanguage{SMTLib}{
morekeywords={forall, exists, and, or, not, ite, Int},
keywordstyle=\color{black!50},
morecomment=[l]{;},
commentstyle=\color{gray}\itshape,
basicstyle=\ttfamily\small,
mathescape=true
}

\newcommand{\mloop}{\mathit{loop}}
\newcommand{\mloopt}{\mathit{loop2}}
\newcommand{\mcompr}{\mathit{compr}}
\newcommand{\mcond}{\mathit{cond}}
\newcommand{\mmod}{\mathit{mod}}
\newcommand{\mathleft}{\@fleqntrue\@mathmargin0pt}
\newcommand{\ite}[3]{\text{if } #1 \text{ then } #2 \text{ else } #3}
\newcommand{\mdiv}{\mathit{div}}
\newcommand{\mmods}{\ \mathit{mod}\ }
\newcommand{\mdivs}{\ \mathit{div}\ }

\usepackage{enumitem}
\setlist[itemize]{nosep, topsep=0pt, partopsep=1pt, leftmargin=*, labelsep=5pt}

\makeatletter
\renewcommand\section{\@startsection{section}{1}{\z@}%
                       {-12\p@ \@plus -4\p@ \@minus -4\p@}%
                       {8\p@ \@plus 4\p@ \@minus 4\p@}%
                       {\normalfont\large\bfseries\boldmath
                        \rightskip=\z@ \@plus 8em\pretolerance=10000 }}
\makeatother

\makeatletter
\renewcommand{\paragraph}[1]{%
  \@startsection{paragraph}{4}{0pt}{3pt}{0pt}{\itshape}%
  {#1}\hspace{1pt}
}
\makeatother

\begin{document}
\title{Learning Conjecturing from Scratch}
\author{Thibault Gauthier\orcidID{0000-0002-7348-0602} \and Josef 
Urban\orcidID{0000-0002-1384-1613}}
\institute{Czech Technical University in Prague\\
	\email{email@thibaultgauthier.fr}, \email{josef.urban@gmail.com}
}
\authorrunning{T. Gauthier, J. Urban}
\titlerunning{Learning Conjecturing from Scratch}
	
	\maketitle

\begin{abstract}
We develop a self-learning approach for %
conjecturing of induction predicates on a dataset of 16197 problems derived from the OEIS.
These problems are hard for today's SMT and ATP systems because they 
require a combination of inductive and arithmetical reasoning.

Starting from scratch, our approach consists of a feedback loop that iterates between
(i) training a neural translator to learn the correspondence between the problems solved so far and the induction predicates useful for them, 
(ii) using the trained neural system to generate many new induction predicates for the problems,
(iii) fast runs of the z3 prover attempting to prove the problems using the generated predicates,
(iv) using heuristics such as predicate size and solution speed on the proved 
problems to choose the best predicates for the next iteration of training.

The algorithm discovers on its own many interesting induction predicates, ultimately solving 5565 problems, compared to 
2265 problems solved by CVC5, Vampire or Z3 in 60 seconds.
\end{abstract}

	\section{Introduction: Induction, OEIS and AI Mathematicians}

Proof by induction is a fundamental tool in mathematics, essential
for reasoning about inductive structures such as trees, lists or 
integers. Efficient automation of induction is a key challenge in automated 
theorem proving (ATP), with direct implications for software verification and 
mathematical research.

A large source of induction problems  is reasoning about
integer sequences.  The Online Encyclopedia of Integer Sequences
(OEIS)~\cite{oeis} covers over 350000 integer sequences from many
domains, often with several proposed explanations for a
sequence. Proving the equivalence of such explanations varies from
easy to very hard. Some of these equivalences are open mathematical
conjectures.

Recently, our program synthesis
AI system~\cite{DBLP:journals/ijar/GauthierOU23} has generated millions
of such explanations, covering more than 130000 OEIS sequences. When a
new algorithm for a particular sequence (e.g. primes) is discovered
automatically by the system, it is only tested on a finite fragment of
the sequence.  One would however also like to automatically prove that
the algorithm is equivalent to the previously known algorithms
(instead of e.g. just computing pseudoprimes).  Having ATPs that can
routinely prove such conjectures would greatly enhance the
capabilities of such autonomous AI mathematicians,\footnote{See
  e.g. \url{https://t.ly/qd626} for a discussion of an AI-invented
  algorithm for $\sqrt[3]{2}$.} allowing e.g. safe optimization of the invented algorithms by replacing slow subroutines with more efficient equivalent ones.

To measure progress in this domain, the OEIS ATP benchmark\footnote{\url{https://github.com/ai4reason/oeis-atp-benchmark}}~\cite{DBLP:conf/lpar/GauthierBJU23} was recently  introduced, consisting of 
SMT~\cite{barrett2010smt} problems of varying difficulty.
Each problem %
involves proving an equivalence of two
programs: a \emph{small} program and a \emph{fast} program, which
generate the same OEIS sequence. The baseline performance of today's
ATP and SMT systems measured in~\cite{DBLP:conf/lpar/GauthierBJU23}
was relatively low: of the 16197 induction problems, unaided CVC5 (the
best system) could prove 504. 

\subsection{Contributions}
In this work we develop a self-learning approach for %
synthesis of instances of second-order induction which are useful for solving the OEIS problems
with the
\zthree~\cite{demoura:2008:zes:1792734.1792766} SMT solver.
We first give a brief overview of the task of AI-based OEIS program
synthesis which motivates our work on predicate synthesis (Section~\ref{sec:oe1}).
This includes the description of the OEIS ATP benchmark
(Section~\ref{sec:benchmark}) which is our problem dataset.
In Section~\ref{sec:programs} we describe the language of the OEIS programs
and explain their translation to the SMT format. This is important for
understanding the kind of the SMT problems we are solving when proving
equivalence of two programs and the inductive predicates that we need
to invent.

Next, in Section~\ref{sec:eval} we introduce methods for \emph{evaluation, selection and minimization} of the invented predicates using \zthree.
This is an important part of the overall system, which also decides how good are our training data and how they evolve.
In Section~\ref{sec:grammar} we design a \emph{suitable language} for the generated inductive predicates. This is used
in Section~\ref{sec:init} 
where we develop an \emph{initial brute-force method} for generation of a set of sufficiently diverse predicates, suitable for starting the self-learning loop.
Here we introduce several \emph{pruning methods} crucial for exploring the large 
space of inductive predicates. This includes semantic evaluation, selection of only the true predicates, fingerprinting to avoid generation of  equivalent predicates, and restricting the language to the most relevant 
functions.%

In Section~\ref{sec:nmt}, we develop our \emph{self-learning loop} which iterates between training a neural machine translation
(NMT) model~\cite{luong2015effective} on previously 
discovered problem/solution pairs, proposing new predicates for all problems, and evaluating them with \zthree . This 
process continuously improves the 
selection of predicates,
which are in turn used for training, after applying our \emph{data-augmentation} methods. %

Finally, Section~\ref{sec:experiment} evaluates the methods in several ways.
Section~\ref{sec:runs} describes several long self-learning runs which ultimately prove \textbf{5565} of the 16179 problems.
This also results in a strong trained NMT+ATP system that proves
5372 problems in at most 48 seconds, and a large dataset of useful induction predicates.
In Section~\ref{sec:comp}
we for a comparison
develop strong baseline methods which use manual
heuristics for induction. The strongest one proves 4064 problems in 10 seconds, and the union of these baseline methods prove 5079 problems.
We show that union of the manual and self-learning approaches solves \textbf{6351} problems, i.e.,
the feedback loop adds 1272 problems to those solved by manual heuristics.
We analyze the
problem-solving and generalization behavior of an interesting invented
predicate in Section~\ref{sec:evolution} and discuss further examples in the Appendix.

\section{From OEIS Programs to Predicate Synthesis}
\label{sec:oe1}
Our approach builds on our method for OEIS program
synthesis~\cite{DBLP:journals/ijar/GauthierOU23}. This is a feedback
loop that interleaves (i) training neural machine translation to learn
the correspondence between sequences and the programs discovered so
far, and (ii) proposing many new programs for each OEIS sequence by
the trained neural machine translator.  Starting from scratch, this loop has so far invented programs for
more than 130,000 OEIS sequences.
This dataset is the foundation of our benchmark, 
consisting of program equalities that we now aim to verify. 

While our previous work focused on synthesizing \emph{programs}, our focus here is on 
synthesizing \emph{predicates} about those programs, which are then used to instantiate a 
second-order induction axiom. This introduces two key 
challenges:
\begin{compactenum}
\item The space of possible predicates is empirically larger than the 
space of programs, making the search more complex. 
\item Evaluating the 
usefulness of predicates using \zthree is much slower
than evaluating programs—taking, on average, 200 milliseconds per predicate 
versus less than 1 millisecond per program. We are thus able to test far 
fewer predicates than 
programs, limiting our ability to explore the broader search space.%
\end{compactenum}

\subsection{The OEIS ATP Benchmark}\label{sec:benchmark}
Our benchmark introduced in~\cite{DBLP:conf/lpar/GauthierBJU23} consists 
of SMT problems of the form $\forall x \in 
\mathbb{N}.\  f_{\textit{Small}_s}(x) = f_{\textit{Fast}_s}(x)$. Here, $s$ is an OEIS sequence,
$\textit{Small}_s$  and $\textit{Fast}_s$ are the smallest and fastest
programs computing $s$ discovered by our self-learning 
system~\cite{DBLP:conf/aaai/GauthierU23,DBLP:journals/ijar/GauthierOU23}, and 
$f_{\textit{Small}_s}$ and 
$f_{\textit{Fast}_s}$ are the SMT functions representing  
the $\textit{Small}_s$ and $\textit{Fast}_s$ programs.
We will usually drop the index $s$ and just write $\textit{Small}, \textit{Fast}, f_\textit{Small}, f_\textit{Fast}$.
For the purpose of this work, we focus on the subset of this benchmark 
consisting of 16197 problems that were heuristically 
determined~\cite{DBLP:conf/lpar/GauthierBJU23} to require 
induction. These SMT problems include
function definitions for the programs $\textit{Small}$ and $\textit{Fast}$ derived 
from their semantic interpretation (see Section~\ref{sec:programs}), and the 
assertion $\exists c\in\mathbb{Z}.\ c \geq 0 \wedge \neg 
(f_\textit{Small}(c) = 
f_\textit{Fast}(c))$. Thus, if these problems are unsatisfiable then the equality 
holds.
\begin{example}\label{ex:one}
The OEIS sequence \href{http://oeis.org/A000217}{A217} %
are the triangular numbers $0, 1, 3, 6, 10, 15 \ldots$ In our benchmark, the $\textit{Small}_{A217}$ program %
is $\mloop(X+Y,X,0)$\footnote{Note that in $\mloop(X+Y,X,0)$ the interpretation 
(Section~\ref{sec:programs}) of the program $X+Y$ will be defined as  $f_{X+Y}$ 
where $f_{X+Y} = \lambda x y. f_{X+Y}(x,y)=\lambda x y. 
(f_X(x,y)+f_Y(x,y))=\lambda x y. x+y$. So the reader can read this program as 
$[\mloop](\lambda x y. x+y,\lambda x y. x,\lambda x y. 0)$ where 
$[\mloop]: (\mathbb{Z}^2
	\mapsto \mathbb{Z})^3 \mapsto  (\mathbb{Z}^2
	\mapsto \mathbb{Z})$ .} corresponding to the mathematical formula 
	$\sum_{k=1}^x k$. 
And our $\textit{Fast}_{A217}$ program %
is $(X \times X + X) \mdivs 2$.
The problem of determining if $\textit{Small}_{A217}$ and $\textit{Fast}_{A217}$ are equal is represented\footnote{The details of the translation will be given in Section~\ref{sec:smt}.}
by the following SMT 
formulas\footnote{\url{https://tinyurl.com/yhx239d9/A217.smt2}} in 
our benchmark.
\begin{lstlisting}[language=SMTLib, caption={}]
(forall ((x Int) (y Int)) (= (f x y) (+ x y)))
(forall ((x Int)) (= (g x) x))
(= h 0)
(forall ((x Int) (y Int)) 
  (= (u x y) (ite (<= x 0) y (f (u (- x 1) y) x))))
(forall ((x Int)) (= (v x) (u (g x) h)))
(forall ((x Int)) (= (small x) (v x)))
(forall ((x Int)) (= (fast x) (div (+ (* x x) x) 2)))
(exists ((c Int)) 
  (and (>= c 0) (not (= (small c) (fast c)))))
\end{lstlisting}
\end{example}

\section{OEIS Programs}\label{sec:programs}
The OEIS programs in our benchmark were produced using a compact grammar, as 
described in~\cite{DBLP:conf/aaai/GauthierU23}. In this section, we 
define the syntax in which these programs were written and provide formal 
semantics. The interpretation functions introduced are used to
convert programs into SMT functions (see Section~\ref{sec:smt}).

\paragraph{Syntax}
The set $\mathbb{P}$ of programs in our language is inductively defined to be 
the smallest set that includes:
\begin{itemize}
\item first-order operators: if $A,B,C \in 
\mathbb{P}$ then \\ \mbox{     } 
$0,1,2,X,Y, A + B, A - B, A \times B, A \mdivs B, A \mmods B,  \mcond(A,B,C) \in 
\mathbb{P}$.
\item second-order operators: if $A,B,C,F,G \in \mathbb{P}$ then
\\ \mbox{     }  $\mloop(F,A,B)$, $\mloopt (F,G,A,B,C)$, 
$\mcompr(F,A) \in 
\mathbb{P}$.
\end{itemize}
A \textit{loop programs} is a program whose top-level operator is
$\mloop$, $\mloopt$ or $\mcompr$.

\paragraph{Semantics}
Each program $P$ is interpreted by a function $f_P: \mathbb{Z}^2
\mapsto \mathbb{Z}$. The interpretation $f_P$
is recursively defined for every program $P$ by:
\begingroup
\allowdisplaybreaks	
\begin{align*}
&f_{0} (x,y) := 0,\ f_{1} (x,y) := 1,\ f_X(x,y) = x,\ f_Y(x,y)=y\\
&f_{A + B} (x,y) := f_A(x,y) + f_B(x,y),\ f_{A - B}(x,y) := f_A(x,y) - 
f_B(x,y)\\
&f_{A \times B} (x,y) := f_A(x,y) \times f_B(x,y),\ f_{A \mdiv B}(x,y) := 
f_A(x,y) \mdivs f_B(x,y)\\
&f_{A \mmod B}(x,y) := f_A(x,y) \mmods f_B(x,y)\\
&f_{\mcond(A,B,C)}(x,y) :=\ \ite{f_A(x,y) \leq 0}{f_B(x,y)}{f_C(x,y)}\\
&f_{\mloop(F,A,B)}(x,y) = u(f_A(x,y),f_B(x,y))\\
&\hspace{5mm} \mbox{where } u(x,y) = \ite{x \leq 0}{y}{f_F(u(x-1,y),x)}\\
&f_{\mloopt(F,G,A,B,C)}(x,y) := u(f_A(x,y),f_B(x,y),f_C(x,y))\\
&\hspace{5mm} \mbox{where } u(x,y,z) = \ite{x \leq 
	0}{y}{f_F(u(x-1,y,z),t(x-1,y,z))}\\
&\hspace{5mm} \mbox{and } t(x,y,z) = \ite{x \leq 
	0}{z}{f_G(u(x-1,y,z),t(x-1,y,z))}\\
&f_{\mcompr(F,A)}(x,y) := u(f_A(x,y))\\
&\hspace{5mm} \mbox{where } t(x) = \ite{f_F(x,0) \leq 0}{x}{t(x+1)}\\
&\hspace{5mm} \mbox{and } u(x) = \ite{x \leq 0}{t(0)}{t(u(x-1) + 1)}
\end{align*}
\endgroup

The functions $\mmod$ (modulo) and $\mdiv$ (integer division) used in these 
definitions follow the semantics of \sml~\cite{harper1986standard}.

\begin{example}
Intuitively, the program $\mloop(2 \times X,X,1)$ produces the sequence
$a_0 = 1, a_n = 2 \times a_{n-1}$. Similarly, the program $\mloop(X + Y,X,0)$ 
computes the sequence $a_0 = 0, a_n = a_{n-1} + n$. Lastly, the program $\mloop(X 
\times Y,Y,X,1,2)$ generates the sequence $(a_n)$ where $(a_0,b_0) = (1,2)$ and
$(a_n,b_n)=(a_{n-1} \times b_{n-1}, b_{n-1})$.
\end{example}

\paragraph{Terminology} We classify interpretation functions into the 
following categories:
\begin{itemize}
	\item \textit{loop functions}: functions corresponding to loop programs.\\
	The function $f_{\mloop(F,A,B)}$ is named $v$, 
	the function $f_{\mloopt(F,G,A,B,C)}$ is named $w$ 
	and $s$ is additional \textit{loop function} associated with 
	$\mloopt(F,G,A,B,C)$.
	\item \textit{argument functions}: functions interpreting arguments of loop
          programs.\\ Given a loop program $\mloop(F,A,B)$, the \emph{update} 
          function $f_F$ is named $f$, the \emph{bound} 
          function $f_A$ is named $g$, and the \emph{initial value} function $f_B$ 
          is named $h$.
	\item \textit{helper functions}: local recursive functions $t$,$u$ 
	within \textit{loop functions}. For $\mloopt$, $u$ computes the first component and $t$ the second component, while in $\mloop$ only $u$ is used.
\end{itemize}

\paragraph{Additional Function $s$}
We include an additional function $s$ for each program $\mloopt(F,G,A,B,C)$ 
in the grammar for induction predicates. The function $s$ is defined as:
\[s(x,y) := t(f_A(x,y),f_B(x,y),f_C(x,y))\]
This function allows proving properties about the second 
component of $\mloopt$ which would otherwise require using the helper
function $t$.\footnote{Later, we will see that this is necessary as helper functions are disallowed in our
initial candidate generation step (Section~\ref{sec:init}).}

\subsection{Example of our SMT translation}\label{sec:smt}

We show our translation to SMT
on the example of the above program $\mloop(X+Y,X,0)$.
To define the toplevel loop function ($f_{\mloop(X+Y,X,0)}$ named $v$ below), we first create SMT definitions for the 
\textit{argument functions} ($f_{X+Y}$, $f_X$, $f_0$, named $f,g,h$ below).
When writing SMT definitions, we omit dummy arguments, thus some of
the functions become unary or nullary:
\begin{lstlisting}[language=SMTLib, caption={}]
(forall ((x Int) (y Int)) (= (f x y) (+ x y)))
(forall ((x Int)) (= (g x) x))
(= h 0)
\end{lstlisting}
From the update function $f$, one can then
construct the helper function $u$ with this recursive definition:
\begin{lstlisting}[language=SMTLib, caption={}]
(forall ((x Int) (y Int)) 
   (= (u x y) (ite (<= x 0) y (f (u (- x 1) y) x))))
\end{lstlisting}
For a given initial value $y$ and a number of iterations (bound) $x$, this 
function $u(x,y)$ corresponds to the recursive sequence:
\[\{u_0,u_1,u_2, \ldots,u_x, \ldots\} = \{y, f(u_0,1),f(u_1,2), \ldots, 
f(u_{x-1},x), \ldots\}\]
For our particular choice of $f$, $u(x,y)= y + \sum_{k=1}^x k$.
To produce our final definition $v$, we compose $u$ with $g$ and $h$:
\begin{lstlisting}[language=SMTLib, caption={}]
(forall ((x Int)) (= (v x) (u (g x) h)))
\end{lstlisting}
For our particular choice of $f$, $g$ and $h$, $v(x)= u(g(x),h) = u(x,0) = 
\sum_{k=1}^x k$.

One can similarly construct SMT definitions
for %
$\mloopt$ and $\mcompr$.
which match
their semantic interpretation. %
We also number these functions based on the serial numbering of the loop constructs.
e.g. $v_0,u_0,f_0,g_0,h_0$,  $w_1,s_1,u_1,t_1,...$.

\paragraph{Example} (\href{https://oeis.org/A108411}{A108411}: Powers of 3 repeated)
This problem presents two ways of computing the powers of 3 repeated twice. The
fast program $w$ stores the value 3 into the second component $s$ of the $\mloopt$
whereas the small program $v$ recomputes $X+X+X$ at every iteration of $\mloop$.

\vspace{-3mm}
\begin{small}
\begin{equation}
\underbrace{\mloop (\overbrace{X + X + X}^{f_0}, \overbrace{X \mdivs 2}^{g_0}, \overbrace{1}^{h_0})}_{v_0}=
\underbrace{\mloopt (X \times Y, Y, X \mdivs 2, 1, 1 + 2)}_{w_1,s_1}
\tag{\href{https://oeis.org/A108411}{A108411}} \label{eq:sf0} 
\end{equation}
\end{small}

\begin{remark}
Since the semantics of modulo ($\mmod$) and 
integer division ($\mdiv$) differs between \sml and SMT, we define
the \sml $\mdiv$ and $\mmod$ in terms of their SMT counterpart in
the preamble of each SMT problem.
\end{remark}

\subsection{Additional Axioms for Trivial Inductions}\label{sec:triv}
We have changed the original problems to be able to prove
more equalities between programs without requiring induction.
The first change is to share definitions of \textit{loop functions} if 
they are syntactically equal, i.e., we now use the same name for them. %
Second, it is often the case that two \textit{loop functions} $v_0$ and $v_1$,
with the same loop operator (e.g $\mloop$), are not syntactically equal 
but their helper functions are. In this case, we add an equation 
stating the equality between the \textit{helper functions} $u_0$ and $u_1$.
Finally, for two loops of the same kind (e.g $\mloop$)
we observe that if the update functions are proven equal (e.g $f_0,f_1$)
then the \textit{helper functions} (e.g. $u_0,u_1$) must be equal.
These implications (\emph{congruences}) are added as extra axioms to our problems
as they cannot be proven without induction. We will see in Section~\ref{sec:experiment} 
that these
simple modifications make more of the benchmark problems solvable by standard methods.

\section{Candidate Selection via Proof Search}\label{sec:eval}
For each problem in our benchmark, we generate a list of predicates 
(called a \textit{candidate}) using initially brute force  (Section~\ref{sec:init})
and then neurally-guided generation (Section~\ref{sec:nmt}).
We describe here how we determine which candidates are
considered a solution for the problem and when
they should be added to our set of training examples. 
Given a predicate $Q$, we produce a first-order axiom (over $\mathbb{Z}$) by instantiating the 
following second-order induction axiom with $Q$:
\begin{small}
\begin{align*}
	\forall P.\ ((\forall y.\ P(0,y)) \wedge (\forall xy.\ P(x,y) \Rightarrow 
	P(x+1,y))) \Rightarrow (\forall xy.\ 0 \leq x \Rightarrow P(x,y)) \tag{Ind} \label{eq:ind}
\end{align*}
\end{small}
The condition $0 \leq x$ is omitted from the induction step, 
seemingly producing a weaker induction axiom. Both versions are in fact equivalent 
as 
can be seen by
instantiating $P$ by $\lambda xy.\ 0 \leq x \Rightarrow Q(x,y)$.
Given a candidate (list of predicates),  %
we create one instance of axiom \ref{eq:ind} for each predicate in the candidate. 
We then append this list of axioms to the SMT problem.
Finally, we call \zthree on the produced problem for 200 milliseconds (an experimental sweet spot).
If \zthree returns a proof, we refer to the candidate as a \textit{solution}
for that problem.

\paragraph{Minimization and Selection of Solutions}
To increase the quality of our training examples (problem/solution pairs) we
perform a minimization step and a selection step before using the new solutions
for training. %
The \emph{minimization step} %
repeatedly calls \zthree with subsets
of the induction instances to see if any of the predicates can be 
removed. The \emph{selection step} consists of keeping only the shortest solution 
discovered so far for each problem in our database of training examples.

\begin{remark}
In some of our experiments, we keep both the shortest and fastest solutions 
for each problem. When looking for fast solutions, the minimization step is
adapted so that a predicate is removed from the solution if a proof can be found
faster without the corresponding axiom for that predicate.
\end{remark}

  \section{Grammar for Induction Predicates}\label{sec:grammar}
  
Induction predicates are built from quantifier-free first-order formulas.
Given a problem $p$, the set of terms $\mathbb{T}$, the set of literals
$\mathbb{L}$ and the set of formulas
$\mathbb{F}$ are defined to the be smallest sets such that:
\begin{itemize}
\item $0,1,2,x,y\in \mathbb{T}$.
\item If $t_1,t_2 \in \mathbb{T}$ then 
$t_1 + t_2,\ t_1 - t_2,\ t_1 \times t_2,\ t_1 \mdivs t_2,\ t_1 \mmods t_2  \in 
\mathbb{T}$.
\item if $t_1,t_2,t_3 \in \mathbb{T}$ then $\ite{t_1 \leq 0}{t_2}{t_3} \in 
\mathbb{T}$.
\item if $f$ is a $n$-ary \textit{function} appearing in $p$ and
$t_1,\ldots,t_n \in \mathbb{T}$ then
$f(t_1,\ldots,t_n) \in \mathbb{T}$.
\item if $t_1,t_2 \in \mathbb{T}$ then 
$t_1 = t_2,\ \neg (t_1 = t_2),\  t_1 \leq t_2, \neg (t_1 \leq t_2) \in 
\mathbb{L}$.
\item if $l_1,l_2 \in \mathbb{L}$ and $f_1,f_2 \in \mathbb{F}$ 
then $l_1,l_2,f_1 \wedge f_2,\ f_1 \Rightarrow f_2 \in \mathbb{F}$.
\end{itemize}
Each formula $f[x,y]$, we associate a predicate $P=\lambda xy.\ f[x,y]$
that will be used to instantiate the induction axiom.
In the rest of this paper, we omit the $\lambda xy.$ part and
use the formula $f[x,y]$ to represent a predicate.

\begin{remark}
Early results indicate that adding quantifiers to the language of 
induction predicates would only help to solve a few more problems
due to the practical difficulty of instantiating those quantifiers. Indeed,
strong induction, which can be obtained by instantiating weak/standard induction
by a quantified formula, subsumes in theory any induction over the $n$ previous 
steps. However, in practice it performs worse than weak/standard induction,
as shown in Section~\ref{sec:experiment}\ .
\end{remark}

\section{Initial Generation of Candidates}\label{sec:init}	
We first explain 
how candidates (lists of predicates) are 
generated in the initial phase.
The predicates are generated using the 
grammar described in Section~\ref{sec:grammar}.
This generation relies on brute-force enumeration of predicates of increasing
sizes. To get interesting solutions, %
we use
the following methods to reduce the %
space.

First, we only generate predicates on functions we deem the most important:
\emph{loop} functions and $s$ functions (Section~\ref{sec:programs}). Second, we discard equivalent terms
if they are equivalent up to arithmetical reasoning.
Third, we discard formulas that do not depend on the induction variable $x$.
Fourth, we test formulas on some values and keep only the formulas that
are true on all values. Thanks to these heuristics,  we were 
able to increase the number of problems solved by these initial candidates
resulting in a large enough pool of examples to train the NMT.

\subsection{Initial Terms}
To avoid generating many versions of the same kind of terms we define 
the following equivalence relation:
\paragraph{Equivalence for terms without \textit{loop subterms}.}
Two terms $t_1$ and $t_2$ are \emph{equivalent} if they evaluate to the 
same value after substituting $(x,y)$ with integers from the set $I=\{(x,y)\ |\ 
0 \leq x < 10, -5 \leq y < 10\}$. After ordering $I$ using the lexicographical 
order, the list of evaluations on $I$
is called a \emph{fingerprint}.
\vspace{-1mm}
\begin{example}
 	 $x$, $x+0$, $x+1-1$ have the same fingerprint.
\end{example}
\paragraph{Equivalence for terms with \textit{loop subterms}.}
We want to find properties of \textit{loop 
functions} therefore we consider \textit{loop 
functions} to be opaque functions. 
In practice, we do so by assigning a random 
fingerprint for %
each \textit{loop subterm}. This fingerprint does 
not correspond to the real evaluation of the subterm but distinguishes it from 
other terms. The computation of fingerprints for terms with loop subterms then proceeds normally.
\vspace{-1mm}
\begin{example} $f(x) + 0$ and $f(x) \times 1$ have the same fingerprint but
$g(x)$ and $f(x)$ do not, even if $f$ and $g$ 
compute the same function. The reason is that \zthree can prove 
that $f(x) + 0 = f(x) \times 1$ but might not be able to prove that
$f(x)=g(x)$ without induction axioms.
\end{example}
We use a bottom-up approach for term generation. We first create all terms of 
size 1, then all the terms of size 2 and so on. During this process, if two 
terms are equivalent, as defined previously, we only keep the smallest one.
We stop generating terms as soon as we get 1024 terms and
we call this set $\mathbb{T}_{init}$.
   
\subsection{From Terms to Candidates} 
We generate candidates from terms in three stages: literals, predicates, 
candidates. 
\paragraph{Literals}
We generate two types of equations and inequations by randomly 
sampling pairs from $\mathbb{T}_{init}$. We sample 250 equations and inequations
that are true on $I$ and 250 equations and inequations that are not always false 
on $I$. This bias favors true literals. The same process is applied to produce 
disequalities and disinequalities forming a set $\mathbb{L}_{init}$ of 1000 
literals.
\paragraph{Predicates} We construct 4000 predicates (implications or 
conjunctions), verified to be true on $I$, by sampling pairs of 
literals from $\mathbb{L}_{init}$. 
\paragraph{Candidates} From these 4000 predicates we sample 1000 ordered subsets 
of size 4 to produce our initial candidate set.

\section{Generating Candidates using NMT}\label{sec:nmt}

So far, we only generated candidates using brute-force enumeration.
However, similar to OEIS program synthesis, conjecturing can be also
naturally formulated as language modelling and sequence-to-sequence
translation tasks. We have recently started to experiment with
recurrent neural networks (RNNs) with attention~\cite{ChorowskiBSCB15} and transformers~\cite{DBLP:conf/nips/VaswaniSPUJGKP17} in
such symbolic settings 
\cite{DBLP:conf/mkm/WangKU18,abs-1911-04873,DBLP:conf/mkm/UrbanJ20,DBLP:journals/ijar/GauthierOU23}.  In 
particular, a
wide \emph{beam search} using relatively small and fast
encoder-decoder neural machine translation (NMT) system works
surprisingly well as the learning and generative component in our
OEIS program synthesis system \cite{DBLP:journals/ijar/GauthierOU23}.

\paragraph{NMT Framework, Hyperparameters, Data, Overall Loop:}
In this work, we train such NMT models to prune the large search space and generate
relevant candidates.
As in \cite{DBLP:journals/ijar/GauthierOU23}, we use the Luong's NMT
framework~\cite{luong17}, with the hyperparameters found
in~\cite{DBLP:conf/mkm/WangKU18}. In more detail, we train a 2-layer bidirectional LSTM
equipped with the ``scaled Luong'' attention and 512 units. Depending
on the experiment and the data size, we train for 2000 -- 14000 steps\footnote{This training takes 15--130 minutes on a low-end GPU (GTX1080 Ti, 12G RAM).}. The inference (generation of candidates)
uses beam width of 240.\footnote{The inference takes about 75 minutes for the 16197 problems on the same GPU.}
A machine learning model, such as the NMT, is trained on a set
of input/output examples. In our setting, the input is a problem and
the output is a solution %
The initial problem/solution pairs are obtained by checking candidates
from the initial generation (Section~\ref{sec:init}). We then train the 
NMT on 
these pairs, use the trained NMT to generate new candidates, which in turn leads to new problem/solution 
pairs after evaluation (Section~\ref{sec:eval}). This process can be iterated
many times
as seen in our experiments (Section~\ref{sec:experiment}). An interesting and important part of this
process are \emph{data augmentation} methods.

\subsection{Training}\label{sec:train}
Since NMT is 
operating on 
tokens, both the problem (pair of programs) and the solution 
are represented as token sequences. In the \textit{split mode} (cf. Section~\ref{sec:inf}), 
each problem-solution pair is 
decomposed into multiple problem-predicate pairs, one for each predicate in the 
solution.
During each iteration of our self-learning loop, we train the NMT %
on the smallest 
solutions found so far. In some runs, we also add the fastest solutions.

\paragraph{Encoding of a Problem}
The two programs are written in prefix notation and 
separated by the $=$ token.
To emphasize the connection between programs and functions, 
we insert after each loop operator a token referring to its loop function.

\paragraph{Encoding of a Solution}
In our settings, a predicate is a quantifier-free formula, thus it is 
straightforward to write its term tree in prefix notation. 
To encode \textit{loop functions}, we use the lowercase letters $a$ to $t$ as 
tokens. This is possible as our problems contain up to 20 loop functions.
Helper functions and argument functions are represented by a token for
their associated loop functions followed by a number.
For example, the loop function $v_0$ is encoded as $a$ since its loop index
is $0$. Its arguments functions $f_0,g_0,h_0$ are encoded as $a\ 0$, $a\ 1$, $a\ 
2$ and its helper function $u_0$ as $a\ 3$.
A solution is encoded by concatenating the encoding of each of its 
predicates sequentially.
\vspace{-0.3mm}
\begin{example}
We show the encoding for the problem $\textit{Small}_{A217}=\textit{Fast}_{A217}$ from Example~\ref{ex:one}.
  Each training example is written on one line, so we separate the problem and 
solution by the token $>$.
Uppercase letters are used for operators that are not loop functions (e.g. 
$A:0,B:1,C:2,D:+,\ldots, O:\ =$).
\vspace{-0.3mm}
\begin{verbatim}
Problem: (loop:v0 (+ x y) x 0) = (div (+ (* x x) x) 2)
Solution: (= (+ (* x x) x) (* 2 (v0 x)))	
Example: J a D K L K A = G D F K K K C > O D F K K K F C a K	
\end{verbatim}
\end{example}

\vspace{-2mm}
\subsubsection{Data Augmentation}
We teach the NMT how to grammatically use functions that do not appear in its 
default training examples. To this end, we either construct extra examples or 
modify existing ones.

\paragraph{Rare Indices for Loop Functions}
Problems with high number of loops are rare in our dataset. Thus, since
our loops are encoded alphabetically in each problem, letter tokens such as
$i$ might never occur in our solutions. Yet, the token $i$ might be
needed to solve a problem involving at least 9 loop functions.
To address this, we sometimes (10\% of the time) shift function 
tokens by a random offset in 
our encoding, resulting in training data that contain such further indices.

\paragraph{Definition Expansion}
During the initial candidate generation phase, we focused solely on producing 
predicates for \textit{loop functions}. However, some induction proofs require 
induction predicates to be expressed in terms of \textit{helper functions}, and 
incorporating \textit{argument functions} can also be beneficial.
Since our initial examples do not include helper or argument functions, we augment 
the training set by generating two additional variations of each problem-solution 
pair. This is done by randomly expanding function definitions within the 
solution (once for the first variation and twice for the second), thereby 
introducing helper and argument functions into the examples.

\subsection{Neural Generation (Inference)}\label{sec:inf}
The potential solutions are generated by running the trained NMT in
its \emph{neural inference mode}, using each problem (i.e., a program
equality) as a separate input. This is analogous to the task of neural
translation of English sentences to their most statistically plausible
French counterparts, but here we learn to translate from problems to
their statistically most plausible induction predicates. The inference
runs as 240-wide \emph{beam search}, yielding for each
problem 240 outputs.

In the \textit{whole mode}, we generate 240 candidates (where each candidate is a list of 
predicates) for each problem. Each candidate is then evaluated by \zthree
individually (see 
Section~\ref{sec:eval}). This allows NMT to synthesize the next 
predicates in the list \emph{conditioned} on the previously generated predicates. %
This typically leads to good coordination and interplay between the predicates in such list~\cite{DBLP:conf/lpar/PiotrowskiU20}.

In the \textit{split mode}, 240 predicates are generated for each problem. These 
predicates are then randomly combined\footnote{In the split mode, we sample 3--12 predicates to
  create a candidate (see Section~\ref{sec:experiment}).} to form 100 candidates which get evaluated by \zthree . The advantage 
of this approach is that it allows generation of candidates with a large 
number of predicates beyond what the NMT could produce in a single %
output.\footnote{While we set the maximum output length of NMT to be 60 here,
  thanks to combining multiple predicates we get solutions of length over 300.} 
However, a drawback of this mode %
is that the NMT generates the predicates independently. As a result, it 
does not tell us which predicates should naturally co-occur within the 
same problem.
This turns out to be important: We eventually synthesize solutions that need as many as 11 predicates, and predicates that consist of as many as 6 conjuncts.

\vspace{-2mm}
\section{Evaluation}~\label{sec:experiment}
\vspace{-12mm}
\subsection{Self-Learning Runs}\label{sec:runs}
\vspace{-1mm}
We have evaluated the system and its properties in
four longer self-learning runs.
These runs build on each other,
modifying some parameters based on the previous runs.
We use a machine with 36 hyperthreading CPU cores, 72 threads, 256GB RAM and 4
GPUs\footnote{CPU: 2 x Intel Xeon Gold 6140 CPU @ 2.30GHz, GPU: 4 x GTX1080 Ti.}
In real
time, the runs have so far taken three months, still solving new
problems. This is similar to our OEIS experiment, where the
self-learning loop has been solving new sequences for over two years
now. The initial data were produced by the \emph{initial (brute-force) run}
(Section~\ref{sec:init}) using 
1000 candidates per problem.
These candidates were evaluated by \zthree in approximately 9 hours resulting in 
2506 solved 
problems.
Figure~\ref{fig:4run} shows the evolution of the number
of solved problems during each of the four self-learning runs.
\pgfplotscreateplotcyclelist{rw}
{solid, mark repeat = 100, mark phase = 500, mark = *, black\\
	solid, mark repeat = 100, mark phase = 1000, mark = square*, black\\}
\begin{center}
\begin{figure}[t]
	\begin{tikzpicture}[scale=1.0]
		\begin{axis}[
			width=0.5*\textwidth,
			height=0.3*\textwidth,
			xmin=0, xmax=275,
			ymin=2500, ymax=5100,
			cycle list name=rw,
			scaled y ticks = false,
			]
			\addplot table[x=gen, y=sol] {gensol1};
		\end{axis}
	\end{tikzpicture}
	\begin{tikzpicture}[scale=1.0]
		\begin{axis}[
			width=0.5*\textwidth,
			height=0.3*\textwidth,
			xmin=0, xmax=63,
			ymin=4900, ymax=5220,
			cycle list name=rw,
			scaled y ticks = false,
			ylabel = {}
			]
			\addplot table[x=gen, y=sol] {gensol2};
		\end{axis}
	\end{tikzpicture}
	\begin{tikzpicture}[scale=1.0]
		\begin{axis}[
			width=0.5*\textwidth,
			height=0.3*\textwidth,
			xmin=0, xmax=41,
			ymin=5160, ymax=5350,
			cycle list name=rw,
			scaled y ticks = false,
			]
			\addplot table[x=gen, y=sol] {gensol3};
		\end{axis}	
              \end{tikzpicture}
              \hfill
	\begin{tikzpicture}[scale=1.0]
		\begin{axis}[
			width=0.5*\textwidth,
			height=0.3*\textwidth,
			xmin=0, xmax=105,
			ymin=5250, ymax=5430,
			cycle list name=rw,
			scaled y ticks = false,
     		ylabel = {\phantom{a}} 
			]
			\addplot table[x=gen, y=sol] {gensol4};
		\end{axis}
		
	\end{tikzpicture}
	\caption{\label{fig:4run} Cumulative number of solved problems after $x$ 
	iterations
	in each run.\\(\textit{Run-1} top-left, \textit{Run-2} top-right, 
	\textit{Run-3} bottom-left, \textit{Run-4} bottom-right)}
\end{figure}
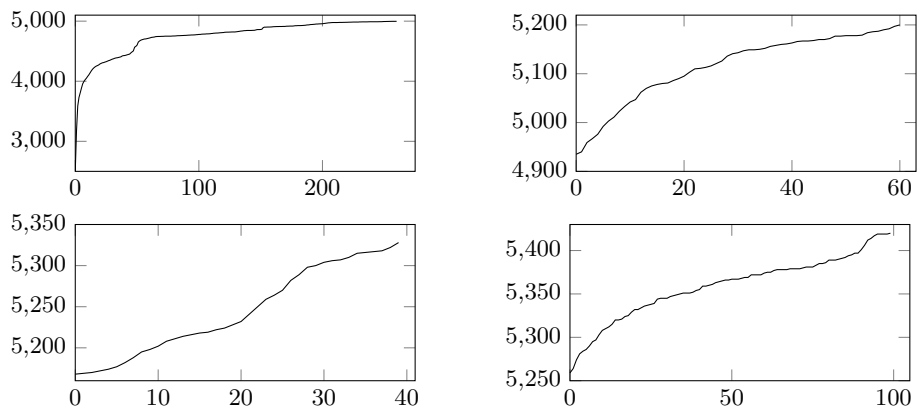
\end{center}
\vspace{-9mm}
\paragraph{Run-1:} was initialized from the 2506 initial solved problems.
The self-learning loop ran here for 260 iterations 
and reached 4997 solved problems.
The learning/generation was performed in the split mode and 
we only trained on the smallest solutions.
Additionally, semantic evaluation was employed to remove
false predicates before producing problems.\footnote{This semantic filter was 
not included in the following runs.}
Each iteration involved 2 to 4 ATP runs, each on outputs produced by differently trained (2000--14000 steps) NMTs.
In each proof attempt we randomly combine $n$ of the 240 generated predicates, where $n$ is randomly chosen (for each run)
from the set $\{1,2,3,4,5,6,8,12\}$.
\paragraph{Run-2:} was initialized from the 4935 problems solved at iteration 187 of 
\textit{Run-1}. This run spanned 60 iterations and reached 5200 solutions.
It again used the split mode but we started to add %
also the fastest solutions to the training data.
We also newly employed data augmentation 
via definition expansions (Section~\ref{sec:train}) to enlarge the training data further,
reaching about 40 thousands examples.
Two ATP runs were done per iteration, using the same approach as in \textit{Run-1}.
\paragraph{Run-3:} was initialized from the 5168 problems solved at iteration 44 of 
\textit{Run-2}. Learning/inference used the whole mode,
i.e., the NMT is here learning and predicting whole candidates instead of 
predicates, and no random combining of predicates is done.
During the learning phase, the NMT was trained on both the fastest and 
smallest solutions discovered but omitting data augmentation 
through definition expansion. The run took 40 iterations, reaching 5329 
solutions.

\paragraph{Run-4:} was initialized from the 5259 problems solved at iteration 23 of \textit{Run-3}.
It used the same parameters as 
\textit{Run-3} with the addition of data augmentation via definition expansion.
Run-4 took 99 iterations, reaching 5420 solutions.

\paragraph{Further Discussion}
The four runs solve in total \textbf{5565} problems. Figure~\ref{fig:4run} shows several sudden increases in the rate at
which problems are solved across the iterations.  These inflections
likely come from the discovery of useful induction patterns, see
Section~\ref{sec:evolution} and Table~\ref{tab:evo1} for an example.  Depending on the quality
of the predictions and other factors, each (parallelized) ATP run within an iteration
takes 1--3 hours. This means that
with the NMT learning/inference we can do about 2--8 iterations
per day.  Further interesting effects include: (i) As the iterations
progress, the ATP runs solve more and more problems, i.e., the NMT
memorizes the discovered solutions well. The record is 5372 problems
solved in iteration 98 of \textit{Run-4}. This means that the final trained NMT+ATP system
outperforms the union of all manual heuristics (Section~\ref{sec:comp}) in a CPU-time comparison.\footnote{The CPU limit in a single run with 240 predictions is at most $240*0.2=48$ seconds. The NMT prediction time is negligible.}
(ii) Solutions of a single problem can evolve a lot both in speed and size.
Our system found 11 short solutions for problem A11914\footnote{\url{https://tinyurl.com/yhx239d9/A11914.smt2}}, which however differ a lot by their speed. One of the longest evolutions occurred for problem A176916,\footnote{\url{https://tinyurl.com/yhx239d9/A176916.smt2}} with 46 different solutions. See Appendix~\ref{app:big} for details.

\vspace{-2mm}
\subsection{Comparison with Manual Heuristics}\label{sec:comp}
We compare our results with \cvc and its induction 
strategy. We also measure how many problems can be solved by \zthree or \cvc
when helped by manually-selected induction axioms. Table~\ref{tab:res1} shows these results.
\begin{table}[t]
	\centering
\begin{tabular}{lcccccccccccc}\toprule
Prover & \multicolumn{10}{c}{Induction over $n$ previous terms} & Str. Ind.& Total 
\\
& 0 & 1 & 2 & 3 & 4 & 5 & 6 & 7 & 8 & 9 &  &  \\
\cmidrule{2-11}
\zthree & 1398 & 1806 & 3461 & 4019 & 4064 & 4049 & 4002 & 3937 & 3847 & 3814 
  & 1403 & 4812\\
\cvc & 1565 & 1742 & 2220 & 1827 & 1642 & 1289 & 1124 & \phantom{0}994 & 
\phantom{0}872 & \phantom{0}788
& 1415 & 2702\\
\bottomrule
\end{tabular}
\caption{\label{tab:res1}Number of problems solved out of the 
	16197 problems
	in our benchmark with a 10 seconds timeout. \cvc was run 
        with its induction strategy.}
\vspace{-4mm}
\end{table}
All these problems contain the additional axioms for trivial inductions from
Section~\ref{sec:triv}.
The first column (0) presents problems with no instances of the induction axiom.
\zthree solves 1398 of those problems which is to be 
compared with the 7 problems \zthree could solve without the axioms for trivial inductions
in \cite{DBLP:conf/lpar/GauthierBJU23}. \zthree and \cvc solve together in 10s 1932 problems.
If we add \vampire with its induction strategy and run it with a 60s time limit, 
the 
systems solve jointly 2265
problems.\footnote{\vampire performs well in 60s (1504 problems) but solves only
  768 problems in 10s. Hence we don't use it for the further 10s evaluations here.}
Columns (1-9) show the numbers of problems
solved when we instantiate the induction axiom by the following family of 
predicates where $a = f_\mathit{Small}$ and $b = f_\mathit{Fast}$:
\begin{align*}
& 0 \leq x \Rightarrow a(x) = b(x) \tag{1} \label{eq:man1}\\
& 0 \leq x \Rightarrow (a(x) = b(x) \wedge a(x+1) = b(x+1))  \tag{2} 
\label{eq:man2}\\
& 0 \leq x \Rightarrow (a(x) = b(x) \wedge a(x+1) = b(x+1) 
\wedge a(x+2) = b(x+2)) \tag{3} \label{eq:man3}\\
& \ldots
\vspace{-2mm}
\end{align*}
The Str. Ind. column refers to the inclusion of the strong induction axiom 
obtained by instantiating the induction axiom used in our experiments with the 
predicate 
$\lambda xy.\ (0 \leq x \Rightarrow (\forall z.\ 
0 \leq z \leq x \Rightarrow a(z) = b(z)))$.

The union of all \zthree and \cvc runs in Table~\ref{tab:res1} solves
4907 problems. If we add also the 60s runs (including \vampire), the
total number of problems solved using these manual heuristics is
5079. This is quite remarkable compared to the previous results
from~\cite{DBLP:conf/lpar/GauthierBJU23}, however still about 500
problems less than the result of our feedback loop. The union of the two approaches solves \textbf{6351} problems, i.e., the feedback loop adds
\textbf{1272} problems to those solved by manual heuristics.

\vspace{-2mm}
\subsection{Example of Influential General Discovery}\label{sec:evolution}

One of the discoveries which produced
the  sudden jump %
at iteration 41 in \textit{Run-1} (Table~\ref{tab:evo1}) is the predicate (where we dropped the loop
indices):
\begin{equation}
s(x) = s(1) \wedge v(1 + x) = w(x) + v(x) + w(x) \tag{P1} \label{eq:pred1}
\end{equation}
This can be used to prove equalities of programs such as those for \href{https://oeis.org/A198766}{A198766}:\footnote{Numbers like $5$ are a short notation for the expressions in our language, e.g. $1+2+2$.}
\vspace{-1mm}
\begin{equation}
\underbrace{\mloop (5 \times (2 + X), X, 1)}_{v} + 2 =
\underbrace{\mloopt (X \times Y, Y, X, 7, 5)}_{w,s} \mdivs 2 \tag{Eq1} 
\label{eq:Eq1}
\end{equation}
Here, the left-hand side loop %
defines a recursive sequence $v(0) = 1$, 
$v(1+x) = 5v(x) + 10 $ . %
The right-hand side loop
defines the sequence %
$(w(x))$ where $(w(0),s(0)) = (7,5), (w(x+1),s(x+1)) = (w(x) \times s(x),s(x))$. 
Thus, a closed form of $w(x)$ is $7 \times 5^x$.
We want to prove \ref{eq:Eq1}, i.e.,  $v(x) + 2 = w(x) \mdivs 2$. 
Assuming the second conjunct of \ref{eq:pred1} holds, we get:
\begin{align*}
&v(1+x)-v(x) = 2w(x) \ 
\underset{v(1+x)}{\Longrightarrow}\  
5v(x) + 10 - v(x) = 2w(x)\\
\Longrightarrow\ &4v(x) + 10 = 2w(x) \underset{div\  4}{\Longrightarrow}\ 
v(x) + (10 \mdivs 4) = w(x) \mdivs 2\\
\Longrightarrow\ &v(x) + 2 = w(x) \mdivs 2 \ \qed  \ \text{(this is Eq1, i.e., 
Q.E.D.)}
\end{align*}
Similarly, the invention of predicate \ref{eq:pred1}  quickly led to proofs of
problems like:
\vspace{-1mm}
\begin{equation}
\underbrace{\mloop (10 \times X + 4, X, 1)}_{v}=
\overbrace{2 \times (\underbrace{\mloopt (X \times Y, Y, X, 2, 10)}_{w} \mdivs 9)}^{w'}
\tag{\href{https://oeis.org/A002278}{A2278}} \label{eq:pred3} 
\end{equation}
stating the equivalence of the sequences $v(0)=1, v(x+1)=10v(x)+4$ and $w'(x) = 
2 \times ((2 \times 10^x) \mdivs 9)$, and
\vspace{-1mm}
\begin{equation}
\underbrace{\mloop (6 \times X + 6, X, 0)}_{v}=
\overbrace{2 \times (\underbrace{\mloopt (X \times Y, Y, X, 3, 6)}_{w} \mdivs 5)}^{w'}
\tag{\href{https://oeis.org/A105281}{A105281}} \label{eq:pred4} 
\end{equation}
stating the equivalence of the sequences
$v(0) = 0, v(x+1)=6v(x) + 6$ and $w'(x) = 2 \times ((3 \times 6^x) \mdivs 5)$.  
An 
interesting property of our invented predicate \ref{eq:pred1}
is its \emph{generality} across these (and further) problems.
It avoids writing explicitly large/varied numbers such as $9$ or
$5$. Indeed, depending on the function $v$, the discrete differentiation operation
$v(x+1)-v(x)$ implicitly present in this predicate reduces to different 
expressions. For instance, in the three mentioned 
examples, we get respectively: $4v(x) + 10$, $9v(x) + 4$ and $5v(x) + 6$. 
Examples involving different kind of patterns in the discovered predicates are 
discussed in Appendix~\ref{app:ex}.
%
%
%
%
%
%
%
%
%
%
%

%
%

%
%
%

%

%
%

%
%
%
%
%
%
%
%
%
%
%
%
%

\begin{table}[t]
	\setlength{\tabcolsep}{8pt}
	\centering
	\begin{tabular}{lrrrrrrrrrr}
		\toprule
		Iteration & 42 & 43 & 44 & 45 & 46 & 47 & 48 & 49 & 50 & 51 \\
		Solutions & 1  & 2  & 5  & 14  & 25 & 27 & 26 & 27 & 27 & 31 \\
		\bottomrule
	\end{tabular}
	\caption{\label{tab:evo1} Number of solutions %
	containing the pattern $v_0 (1 + x) = w_1 (x) + v_0 (x) + \_$
    at iterations 42--51 of \textit{Run-1}.}
\vspace{-4mm}
\end{table}

\vspace{-2mm}
\section{Related Work}%
Automating inductive reasoning has been a long-standing challenge in theorem 
proving and software verification. Our approach is related to prior work on
interactive and automated theorem provers~\cite{DBLP:books/el/RobinsonV01} and especially their AI/ML extensions~\cite{BlaauwbroekCGJK24},
and also takes inspiration from other self-learning 
systems.

\paragraph{Inductive Automated and Interactive Theorem Proving}
Inductive reasoning is a key feature in ATPs, enabling proofs about 
recursive structures and arithmetic properties. \vampire has introduced induction 
techniques~\cite{DBLP:conf/paar/HajduKRV22}, and in particular integer 
induction schemes~\cite{DBLP:conf/cade/HozzovaKV21},
that extend its first-order logic capabilities. 
Similarly, \cvc has 
incorporated strategies for reasoning about inductive 
properties~\cite{reynolds-vmcai15}. 
Several ITPs employ techniques specifically designed for inductive proofs. 
For 
instance, \isabellehol~\cite{isabelle} and \coq~\cite{coq} integrate inductive 
reasoning as a fundamental mechanism, with several recent systems further automating induction in \isabellehol ~\cite{DBLP:conf/ijcai/Nagashima21,NagashimaXWGB23}.
Rippling~\cite{DBLP:journals/ai/BundySHIS93,DBLP:conf/birthday/JohanssonDB10} and 
waterfall strategies~\cite{DBLP:conf/aisc/JiangPF18} have 
been developed to systematically guide proof search by modifying goal expressions. 
Additionally, QuickSpeck~\cite{DBLP:conf/tap/ClaessenSH10,DBLP:conf/ijcar/EinarsdottirHJSS24} and 
HipSpec~\cite{DBLP:conf/cade/ClaessenJRS12} 
have explored automated 
conjecturing 
for equational reasoning to prove properties of programs.

\paragraph{Self Learning}
A directly related self-learning system motivating the work done here is QSynt~\cite{DBLP:conf/aaai/GauthierU23,DBLP:journals/ijar/GauthierOU23}, which
however only tests the invented programs.  AI/TP metasystems such as
MaLARea~\cite{malarea07,us+08}, ATPBoost~\cite{PiotrowskiU18}, BliStr~\cite{blistr} and Grackle~\cite{DBLP:conf/mkm/HulaJJK22} are
using self-learning similar to ours over large formal mathematical
libraries in the context of premise selection and strategy invention. Similar self-learning
has been recently introduced for internal guidance of ATPs such as
rlCoP~\cite{DBLP:conf/nips/KaliszykUMO18},
E/ENIGMA~\cite{DBLP:conf/itp/JakubuvU19},
Vampire/Deepire~\cite{DBLP:conf/frocos/Suda21} and
iProver~\cite{DBLP:conf/lpar/ChvalovskyKPU23}, as well as for direct
neural term synthesis in instantiation
proving~\cite{DBLP:journals/jsc/PiepenbrockUKOHJ25}.
In AlphaGoZero~\cite{silver2017mastering}, a self-learning model learns 
to play %
Go from scratch, improving solely through self-play and 
feedback. DreamCoder~\cite{DBLP:conf/pldi/EllisWNSMHCST21}, a self-learning system 
for program synthesis, illustrates how a mix of %
learning and domain specific algorithms can be used to construct program 
abstractions from examples. 
Systems like AlphaProof~\cite{alphaproof} use self-learning to address International Mathematical Olympiad problems.

\vspace{-2mm}	
\section{Conclusion and Future Work}
In this work, we have added instances of the second-order induction axiom to 
the \zthree prover to prove equalities between OEIS programs.
These instances were generated using a self-learning process, ultimately allowing \zthree to prove 5565 of the 16197 problems. 
We compared our results to state-of-the-art provers for induction and 
arithmetical reasoning, also with heuristically added induction instances.
Our approach outcompetes these methods by about 500 problems, adding 1272 solutions. The union of all these approaches solves 6351 problems, which is a considerable improvement over the previous state of the art in~\cite{DBLP:conf/lpar/GauthierBJU23}.
The byproducts of this work are a strong trained NMT+ATP system that proves 5372 problems in at most 48 seconds, and a large dataset of useful induction predicates. These can be trained on and further analyzed by humans, which we have started to do here, showing
that some of them are interesting and general. 

In the future, we would like to scale our approach to 
much larger sets of problems such as the millions of programs coming from our 
latest OEIS synthesis run. We hope that our methodology can be applied to
related problems,
such as verification of loop 
invariants in industrial programs which are typically much longer than the
OEIS programs. Finally, we
would like to %
apply the approach to
other forms 
of conjecturing such as generating instantiations of first-order 
quantifiers and synthesizing intermediate lemmas (cuts) to simplify large proofs.

	\bibliographystyle{plain}
	\bibliography{biblio}
	
\newpage
\appendix

\section{Examples}\label{app:ex}
We present a selection of example problems found during the 
self-learning runs but not found when manually adding induction predicates.

\paragraph{Example 1} (\href{https://oeis.org/A002411}{A2411}: Pentagonal 
pyramidal numbers)\\
This problem is a variation of the famous Gaussian sum.
\begin{align*}
	&\mbox{Problem: }\underbrace{\mloop(X+Y,X,0)}_{v} \times X = (((X \times X) + 
	X) 
	\mdivs 2) \times X\\
	&\mbox{Math: } x \times \sum_{k=1}^x{k}  = \frac{x \times x + x}{2} 
	\times 
	x\ \ \ \mbox{Solution: }0 \leq x \wedge x \times x + x = 2 \times v(x)
\end{align*}
The proof requires one predicate on the \textit{loop function} $v$ defining
$\mloop(X+Y,X,0)$. We can also observe that our machine learning algorithm 
restricted the induction step
to nonnegative integers by adding the conjunct $0 \leq x$.

\paragraph{Example 2} (\href{http://oeis.org/A059826}{A59826})\\
This problem is another summation problem that additionally requires a 
generalization step as a direct induction on the problem equality would fail.
\begin{align*}
	&\mbox{Problem: }\mloop(X+Y+Y,X \times X, 1) = 1 + (1 + X \times X) 
	\times(X \times X)\\
	&\mbox{Math: } 1 + \sum_{k=1}^{x \times x}2k = x^4 + x^2 + 1\ \ \ 
	\mbox{Solution: 
	} 
	u(x,1) - 1 = x \times x + x \tag{P2} \label{eq:p2}
\end{align*}
The function $u$ is the helper function for the $\mloop(X+Y+Y,X \times 
X,1)$. Thus, $u(x,y) = y + \sum_{k=1}^x 2k$ and $u(x,1) = 1 + \sum_{k=1}^x 2k 
= 1 + (x \times x + x)$.
Our system conjectured the general lemma~\ref{eq:p2} and indeed proved it by 
induction.
~\ref{eq:p2} can then be instantiated
by $x \times x$ to prove the problem. There is no easy way
how to prove the problem directly by induction over $n$ previous steps where $n$ 
is fixed.

\paragraph{Example 3} (\href{http://oeis.org/A001026}{A1026}: powers of 17)
This problem presents two ways of computing the powers of 17. The
fast program $w$ saves time by storing the value of $17=1 + (2 \times (2 \times (2 
+ 2)))$ into the second component $s$ of $\mloopt$
whereas the small program $v$ recomputes 17 at every iteration of $\mloop$.
The value 17 is computed as $1 + (2 \times (2 \times (2 + 2)))$ in the fast 
program
whereas in the small program the update
function $f$ defining $\mloop (X \times X, 2, 2) \times X + X$ can be simplified 
to $16 \times 
X + X = 17 \times X$.
\begin{align*}
	&\mbox{Problem: }
	\underbrace{\mloop (17 \times X, X, 1)}_{v} = 
	\underbrace{\mloopt (X \times Y, Y ,X, 1, 17)}_{w,s}\\
	&\mbox{Math: } v(x) \mbox{ and } w(x) \mbox{ are equivalent, where } v(0)=1, 
	v(x+1)=17  \times v(x) \\ & \mbox{   \ \ \ \ \ \      and } (w(0),s(0)) = 
	(1,17), 
	(w(x+1),s(x+1)) = (w(x) \times s(x),s(x)) \\
	&\mbox{Solution: }v(x) = w(x) \wedge s(x) = v(1)
\end{align*}
In the solution, the function $v$ corresponds to the outermost $\mloop$ 
of the left-hand side and the functions $w$ and $s$ correspond to the
$\mloopt$ program on the right-hand side. 
The second conjunct states $s$ is a constant function. This fact is helpful
to inductively prove the first conjunct expressing the equality between the two 
loop functions $v$ and $w$.

\paragraph{Example 4} (\href{http://oeis.org/A205646}{A205646}: empty faces in 
Freij's family of Hansen 
polytopes)\\
We first simplify the original problem by evaluating constant programs
and reducing polynomials. The problem 
can then be restated as:
\[\underbrace{\mloop(3 \times X, X, 1)}_{v} + 16 = 
\underbrace{\mloopt(X \times Y, Y, X 
	\mdivs 2, \mloop 
	(3,x \mmods 2,1), 9))}_{w,s} + 16\]
Mathematically, the equation is $3^x + 16 = 3^{(x \mmods 2)} \times 9^{(x \mdivs 
	2)} + 16$. The fast program is saving time by computing $3^x$ as $3^{(x \mmods 
	2)}\times 9^{(x \mdivs 2)} $  (this is the basis of the \emph{fast 
	exponentiation} 
algorithm~\cite{DBLP:reference/crc/2005ehcc}). 
Similarly to the previous example, it also stores 
the value of $9=1 + 2 \times (2 + 2)$ in the second component of the loop $y$ to 
avoid recomputing it at 
each iteration of the loop.
The solution consists of 6 predicates: 
\begin{align*}
	&w (1 + x) = v (1 + x) \wedge w(x) = v(x), v(x) = w(x) \wedge w (1 + x) = v (1 
	+ 
	x)\\
	&x \leq w(2), x \leq w(x), s(x) = s(1), s (x + 1) = s(x)
\end{align*}
The first two predicates (which are equivalent) allow \zthree to prove the 
statement by doing an induction over the two previous steps, which is necessary
and thus \zthree can perform case splitting on whether the number of iteration is 
even or odd. The last two predicates state that $s$ is constant. We are not sure 
of the purpose of the predicates $x \leq 
w(2)$ and $x \leq w(x)$.
Yet, we know that \zthree is not able to find a proof 
when these predicates are removed because
the minimization step would have removed them already.

\section{Spectrum of Solutions}\label{app:big}
We show the speeds (in numbers of perf instructions) and the corresponding lists of induction predicates that allow \zthree to solve two problems.

\paragraph{11 Solutions of A11914:}
Our problem for \href{https://oeis.org/A011914}{A11914} is as follows:
\begin{small}
\begin{verbatim}
(((loop:v0 ((x - 2) + y) x 1) * x) div (loop:v1 (x * x) 2 2)) 
= (((1 - x) * ((2 - x) * x)) div (2 * (2 * (2 * (2 + 2)))))
\end{verbatim}
\end{small}
Below we show the solutions, sorted by speed. Note that all of them are
quite short but there are considerable jumps between the solution
times. For, example, inventing the addition of the conjunct \texttt{(<= 0 x)}
produced major speedups. This is likely an artefact of our induction
scheme lacking this precondition explicitly in its antecedent.

\begin{small}
\begin{verbatim}
103430536 (~ (<= y (u1 0 h1))) | (/\ (==> (<= 0 x) (= (+ x (* x x)) 
                       (* 2 (- (u0 x 1) (- 1 (* 2 x)))))) (<= 0 x))
113177652 (<= z (u1 0 h1)) | (/\ (= (+ x (* x x)) 
                        (* 2 (- (u0 x 1) (- 1 (* 2 x))))) (<= 0 x))
113237981 (<= y (u1 0 h1)) | (==> (<= 0 x) (/\ (= (+ x (* x x)) 
                       (* 2 (- (u0 x 1) (- 1 (* 2 x))))) (<= 0 x)))
166979007 (/\ (= (+ x (* x x)) (* 2 (- (u0 x 1) (- 1 (* 2 x))))) (<= 0 x)) 
          | (<= x (u0 2 1))
181398047 (==> (<= 0 x) (==> (<= 0 x) (= (+ (* x x) x) 
                                       (* 2 (- (u0 x 1) (- 1 (* 2 x)))))))
190537593 (==> (<= 0 x) (= (+ (* x x) x) (* 2 (- (u0 x 1) (- 1 (* 2 x))))))
199580137 (/\ (<= 0 x) (= (+ x (* x x)) (* 2 (- (u0 x 1) (- 1 (* 2 x))))))
253312197 (/\ (= (+ x (* x x)) (* 2 (- (u0 x 1) (- 1 (* 2 x))))) (<= 0 x))
498820817 (= (+ (* x x) x) (* 2 (- (u0 x 1) (- 1 (* x 2)))))
786879090 (= (+ (* x x) x) (* 2 (- (u0 x 1) (- 1 (+ x x)))))
2166828683 (= (+ (* x x) x) (* 2 (- (u0 x 1) (- 1 (* 2 x)))))
\end{verbatim}
\end{small}

\paragraph{46 Solutions of A176916:}
Our problem for \href{https://oeis.org/A176916}{A176916} is as follows:
\begin{small}
\begin{verbatim}
(1 + (((2 * (x + x)) + (loop:v0 ((2 * (x + x)) + x) x 1)) + x)) = 
((1 + (loop2:w1 (x * y) y (x div 2) (loop:v2 (1 + (2 + 2)) (x mod 2) 1) 
       (loop:v3 (x * x) 1 (1 + (2 + 2))))) + ((2 * (x + x)) + x))
\end{verbatim}
\end{small}
Below we show the solutions, sorted by speed. Note that in this case the fastest
solution is shorter than some of the slower solutions -- the result of the
gradual optimization and propagation of related solutions in the
self-learning loop.
\begin{tiny}
\begin{verbatim}
36553504 (/\ (/\ (<= 0 x) (= (w1 (+ 1 x)) (u0 (+ 1 x) h0))) (==> (<= 0 x) (/\ (= (w1 x) (u0 x 1)) 
         (/\ (= j1 (s1 (+ 1 x))) (= (s1 x) (s1 (+ 1 x))))))) | (<= (+ x 0) (f0 (u2 1 1)))
36571395 (/\ (/\ (<= 0 x) (= (w1 (+ 1 x)) (u0 (+ 1 x) h0))) (==> (<= 0 x) (/\ (= (w1 x) (u0 x 1)) 
         (/\ (= j1 (s1 (+ 1 x))) (= (s1 x) (s1 (+ 1 x))))))) | (<= (- x 0) (f0 (u2 1 1)))
36589425 (/\ (/\ (/\ (<= 0 x) (= (w1 (+ 1 x)) (u0 (+ 1 x) h0))) (<= 0 x)) (==> (<= 0 x) (/\ (= (w1 x) 
         (u0 x 1)) (/\ (= j1 (s1 (+ 1 x))) (= (s1 x) (s1 (+ 1 x))))))) | (<= (+ x 0) (f0 (u2 1 1)))
36603131 (/\ (/\ (/\ (<= 0 x) (= (w1 (+ 1 x)) (u0 (+ 1 x) h0))) (<= 0 x)) (==> (<= 0 x) (/\ (= (w1 x) 
         (u0 x 1)) (/\ (= j1 (s1 (+ 1 x))) (= (s1 x) (s1 (+ 1 x))))))) | (<= (- x 0) (f0 (u2 1 1)))
36609511 (/\ (/\ (/\ (<= 0 x) (= (w1 (+ 1 x)) (u0 (+ 1 x) h0))) (<= 0 x)) (==> (<= 0 x) (/\ (= (w1 x) 
         (u0 x 1)) (/\ (= j1 (s1 (+ 1 x))) (= (s1 x) (s1 (+ 1 x))))))) | (<= x (ite (<= 1 0) 0 (f0 (u2 1 1))))
36611919 (/\ (/\ (= (w1 (+ 1 x)) (u0 (+ 1 x) h0)) (<= 0 x)) (==> (<= 0 x) (/\ (= (w1 x) (u0 x 1)) 
         (/\ (= j1 (s1 (+ 1 x))) (= (s1 x) (s1 (+ 1 x))))))) | (<= x (ite (<= 1 0) 0 (f0 (u2 1 1))))
36898605 (/\ (/\ (/\ (= (w1 (+ 1 x)) (v0 (+ 1 x))) (<= 0 x)) (= (w1 x) (u0 (g0 x) 1))) (/\ (= j1 (s1 (+ 1 x))) 
         (= (s1 x) (s1 (+ 1 x))))) | (<= x (ite (<= 1 0) 0 (f0 (u2 1 1))))
36899304 (/\ (/\ (= (w1 (+ 1 x)) (v0 (+ 1 x))) (<= 0 x)) (/\ (/\ (= (w1 x) (u0 (g0 x) 1)) (= j1 (s1 (+ 1 x)))) 
         (= (s1 x) (s1 (+ 1 x))))) | (<= x (ite (<= 1 0) 1 (f0 (u2 1 1))))
36899334 (/\ (/\ (/\ (= (w1 (+ 1 x)) (v0 (+ 1 x))) (<= 0 x)) (= (w1 x) (u0 (g0 x) 1))) (/\ (= j1 (s1 (+ 1 x))) 
         (= (s1 x) (s1 (+ 1 x))))) | (<= x (ite (<= 2 0) 0 (f0 (u2 1 1))))
36899880 (/\ (/\ (/\ (= (w1 (+ 1 x)) (v0 (+ 1 x))) (<= 0 x)) (= (w1 x) (u0 (g0 x) 1))) (/\ (= j1 (s1 (+ 1 x))) 
         (= (s1 x) (s1 (+ 1 x))))) | (<= x (ite (<= 1 0) 1 (f0 (u2 1 1))))
36900308 (/\ (/\ (= (w1 (+ 1 x)) (v0 (+ 1 x))) (<= 0 x)) (/\ (= (w1 x) (u0 (g0 x) 1)) (/\ (= j1 (s1 (+ 1 x))) 
         (= (s1 x) (s1 (+ 1 x)))))) | (<= x (ite (<= 1 0) 1 (f0 (u2 1 1))))
36900781 (/\ (/\ (= (w1 (+ 1 x)) (v0 (+ 1 x))) (/\ (<= 0 x) (= (w1 x) (u0 (g0 x) 1)))) (/\ (= j1 (s1 (+ 1 x))) 
         (= (s1 x) (s1 (+ 1 x))))) | (<= x (ite (<= 1 0) 1 (f0 (u2 1 1))))
36901019 (/\ (= (w1 (+ 1 x)) (v0 (+ 1 x))) (/\ (<= 0 x) (/\ (= (w1 x) (u0 (g0 x) 1)) (/\ (= j1 (s1 (+ 1 x))) 
         (= (s1 x) (s1 (+ 1 x))))))) | (<= x (ite (<= 1 0) 1 (f0 (u2 1 1))))
36901696 (/\ (= (w1 (+ 1 x)) (v0 (+ 1 x))) (/\ (<= 0 x) (/\ (= (w1 x) (u0 (g0 x) 1)) (/\ (= j1 (s1 (+ 1 x))) 
         (= (s1 x) (s1 (+ 1 x))))))) | (<= x (ite (<= 1 0) 2 (f0 (u2 1 1))))
36905901 (/\ (= (w1 (+ 1 x)) (v0 (+ 1 x))) (/\ (<= 0 x) (/\ (= (w1 x) (u0 (g0 x) 1)) (/\ (= j1 (s1 (+ 1 x))) 
         (= (s1 (+ 1 x)) (s1 x)))))) | (<= x (ite (<= 1 0) 2 (f0 (u2 1 1))))
37666884 (/\ (/\ (<= 0 x) (= (w1 (+ 1 x)) (u0 (+ 1 x) h0))) (/\ (= (w1 x) (u0 x 1)) (/\ (= j1 (s1 x)) 
         (= (s1 (+ 1 x)) (s1 x))))) | (= (- x 1) (ite (<= (+ 1 x) 0) (+ 1 (+ y 2)) (f0 (u2 (- 2 1) 1))))
37673659 (/\ (/\ (= (w1 (+ 1 x)) (u0 (+ 1 x) h0)) (/\ (<= 0 x) (= (w1 x) (u0 x 1)))) (/\ (= j1 (s1 x)) 
         (= (s1 (+ 1 x)) (s1 x)))) | (= (- x 1) (ite (<= (+ 1 x) 0) (+ 1 (+ y 2)) (f0 (u2 (- 2 1) 1))))
37894449 (/\ (/\ (= (w1 (+ 1 x)) (u0 (+ 1 x) h0)) (/\ (<= 0 x) (= (w1 x) (u0 x 1)))) (/\ (= j1 (s1 x)) 
         (= (s1 (+ 1 x)) (s1 x)))) | (= (- x 1) (ite (<= (+ 1 x) 0) 0 (f0 (u2 (- 2 1) 1))))
38234573 (/\ (/\ (= (w1 (+ 1 x)) (u0 (+ 1 x) h0)) (/\ (<= 0 x) (= (w1 x) (u0 x 1)))) (/\ (= j1 (s1 x)) 
         (= (s1 (+ 1 x)) (s1 x)))) | (<= (- x 1) (ite (<= y 0) (+ 1 2) (f0 (u2 (- 2 1) 1))))
38235909 (/\ (/\ (/\ (= (w1 (+ 1 x)) (u0 (+ 1 x) h0)) (/\ (<= 0 x) (= (w1 x) (u0 x 1)))) (= j1 (s1 x))) 
         (= (s1 (+ 1 x)) (s1 x))) | (<= (- x 1) (ite (<= y 0) (+ 1 2) (f0 (u2 (- 2 1) 1))))
39173013 (/\ (/\ (/\ (= (w1 (+ 1 x)) (u0 (+ 1 x) h0)) (/\ (<= 0 x) (= (w1 x) (u0 x 1)))) (= j1 (s1 x))) 
         (= (s1 (+ 1 x)) (s1 x))) | (<= x (ite (<= 1 0) (+ 1 (+ 2 2)) (f0 (u2 (- 2 1) 1))))
39184040 (/\ (/\ (/\ (= (w1 (+ 1 x)) (u0 (+ 1 x) h0)) (/\ (<= 0 x) (= (w1 x) (u0 x 1)))) (= j1 (s1 x))) 
         (= (s1 (+ 1 x)) (s1 x))) | (<= x (ite (<= 1 0) (+ 1 2) (f0 (u2 (- 2 1) 1))))
39283153 (/\ (= (w1 (+ 1 x)) (u0 (+ 1 x) 1)) (/\ (/\ (<= 0 x) (/\ (= (w1 x) (u0 x 1)) (= j1 (s1 x)))) 
         (= (s1 (+ 1 x)) (s1 x)))) | (<= x (ite (<= 1 0) (+ 1 1) (f0 (u2 (- 2 1) 1))))
42970322 (= (v1 (divf x 2) (u2 (modf x 2) 1) v3) (ite (<= 2 0) 0 (f3 (ite (<= (- 1 1) 0) (+ 1 (+ 2 2)) (- 1 1))))) | 
         (/\ (= (w1 (+ 1 x)) (u0 (+ 1 x) 1)) (= (w1 x) (u0 x 1))) | (= (ite (<= (divf (+ 1 x) 2) 0) v3 (v1 (- (divf (+ 1 x) 2) 1) 
         (u2 (modf (+ 1 x) 2) 1) (u3 g3 h3))) (v1 (divf x 2) (u2 (modf x 2) 1) (u3 g3 h3)))
43056494 (= (v1 (divf x 2) (u2 (modf x 2) 1) v3) (ite (<= 2 0) 0 (f3 (ite (<= (- 1 1) 0) (+ 1 (+ 2 2)) (- 1 1))))) | 
         (/\ (= (w1 (+ 1 x)) (u0 (+ 1 x) 1)) (= (w1 x) (u0 x 1))) | (= (ite (<= (divf (+ 1 x) 2) 0) j1 (v1 (- (divf (+ 1 x) 2) 1) 
         (u2 (modf (+ 1 x) 2) 1) (u3 g3 h3))) (v1 (divf x 2) (u2 (modf x 2) 1) (u3 g3 h3)))
43117995 (= (v1 (divf x 2) (u2 (modf x 2) 1) v3) (ite (<= 1 0) (+ 1 (+ 2 2)) (f3 (ite (<= (- 1 1) 0) (+ 1 (+ 2 2)) 
         (- (- 1 1) 1))))) | (/\ (= (w1 (+ 1 x)) (u0 (+ 1 x) 1)) (= (w1 x) (u0 x 1))) | (= (ite (<= (divf (+ 1 x) 2) 0) j1 
         (v1 (- (divf (+ 1 x) 2) 1) (u2 (modf (+ 1 x) 2) 1) (u3 g3 h3))) (v1 (divf x 2) (u2 (modf x 2) 1) (u3 g3 h3)))
43178285 (= (v1 (divf x 2) (u2 (modf x 2) 1) v3) (ite (<= 1 0) (+ 1 (+ 2 2)) (f3 (ite (<= (- 1 1) 0) (+ 1 (+ 2 2)) (- 1 1))))) | 
         (/\ (= (w1 (+ 1 x)) (u0 (+ 1 x) h0)) (= (w1 x) (u0 x 1))) | (= (ite (<= (divf (+ 1 x) 2) 0) j1 (v1 (- (divf (+ 1 x) 2) 1) 
         (u2 (modf (+ 1 x) 2) 1) (u3 g3 h3))) (v1 (divf x 2) (u2 (modf x 2) 1) (u3 g3 h3)))
44640560 (/\ (= (w1 (+ 1 x)) (u0 (+ 1 x) 1)) (= (w1 x) (u0 x 1))) | (= (v1 (divf x 2) (u2 (modf x 2) 1) v3) (ite (<= 1 0) 
         (+ 1 (+ 2 2)) (f3 (ite (<= (- 1 1) 0) (+ 1 (+ 2 2)) (- 1 1))))) | (<= y (ite (<= 1 0) (+ 1 (+ 2 2)) (f3 (- 1 1)))) | 
         (= (ite (<= (divf (+ 1 x) 2) 0) j1 (v1 (- (divf (+ 1 x) 2) 1) (u2 (modf (+ 1 x) 2) 1) (u3 g3 h3))) (v1 (divf x 2) 
         (u2 (modf x 2) 1) (u3 g3 h3)))
45818579 (/\ (= (w1 (+ 1 x)) (u0 (+ 1 x) 1)) (= (- 2 (u0 x 1)) (- 2 (w1 x)))) | (= (v1 (divf x 2) (u2 (modf x 2) 1) v3) 
         (ite (<= 1 0) (+ 1 (+ 2 2)) (f3 (ite (<= (- 1 1) 0) (+ 1 (+ 2 2)) (- 1 1))))) | (<= y (ite (<= 1 0) (+ 1 (+ 2 2)) 
         (f3 (- 1 1)))) | (= (ite (<= (divf (+ 1 x) 2) 0) j1 (v1 (- (divf (+ 1 x) 2) 1) (u2 (modf (+ 1 x) 2) 1) v3)) 
         (v1 (divf x 2) (u2 (modf x 2) 1) v3))
45854224 (/\ (= (w1 (+ 1 x)) (u0 (+ 1 x) 1)) (= (- 2 (u0 x 1)) (- 2 (w1 x)))) | (= (v1 (divf x 2) (u2 (modf x 2) 1) v3) 
         (ite (<= 1 0) (+ 1 (+ 2 2)) (f3 (ite (<= (- 1 1) 0) (+ 1 (+ 2 2)) (- (- 1 1) 1))))) | (<= y (ite (<= 1 0) (+ 1 (+ 2 2)) 
         (f3 (- 1 1)))) | (= (ite (<= (divf (+ 1 x) 2) 0) j1 (v1 (- (divf (+ 1 x) 2) 1) (u2 (modf (+ 1 x) 2) 1) j1)) 
         (v1 (divf x 2) (u2 (modf x 2) 1) v3))
45995389 (/\ (= (w1 (+ 1 x)) (u0 (+ 1 x) 1)) (= (- 2 (u0 x 1)) (- 2 (w1 x)))) | (= (v1 (divf x 2) (u2 (modf x 2) 1) v3) 
         (ite (<= 1 0) (+ 1 (+ 2 2)) (f3 (ite (<= (- 1 1) 0) (+ 1 (+ 2 2)) (- (- 1 1) 1))))) | (<= y (ite (<= 1 0) (+ 1 (+ 2 2)) 
         (f3 (ite (<= (- 1 1) 0) (+ 1 (+ 1 2)) (- (- 1 1) 1))))) | (= (ite (<= (divf (+ 1 x) 2) 0) j1 (v1 (- (divf (+ 1 x) 2) 1) 
         (u2 (modf (+ 1 x) 2) 1) j1)) (v1 (divf x 2) (u2 (modf x 2) 1) v3))
47105658 (/\ (= (w1 (+ 1 x)) (u0 (+ 1 x) 1)) (= (- 2 (u0 x 1)) (- 2 (w1 x)))) | (= (v1 (divf x 2) (u2 (modf x 2) 1) v3) 
         (ite (<= 1 0) (+ 1 (+ 2 2)) (f3 (ite (<= (- 1 1) 0) (+ 1 (+ 2 2)) (- (- 1 1) 1))))) | (= (ite (<= (divf (+ 1 x) 2) 0) j1 
         (v1 (- (divf (+ 1 x) 2) 1) (u2 (modf (+ 1 x) 2) 1) j1)) (v1 (divf x 2) (u2 (modf x 2) 1) v3))
47114660 (/\ (= (w1 (+ 1 x)) (u0 (+ 1 x) 1)) (= (- 2 (u0 x 1)) (- 2 (w1 x)))) | (= (v1 (divf x 2) (u2 (modf x 2) 1) v3) 
         (ite (<= 1 0) (+ 1 (+ 2 2)) (f3 (ite (<= (- 1 1) 0) (+ 1 (+ 2 2)) (f3 (- (- 1 1) 1)))))) | (= (ite (<= (divf (+ 1 x) 2) 0) 
         j1 (v1 (- (divf (+ 1 x) 2) 1) (u2 (modf (+ 1 x) 2) 1) j1)) (v1 (divf x 2) (u2 (modf x 2) 1) v3))
47164415 (/\ (= (w1 (+ 1 x)) (u0 (+ 1 x) 1)) (= (- 2 (u0 x 1)) (- 2 (w1 x)))) | (= (v1 (divf x 2) (u2 (modf x 2) 1) v3) 
         (ite (<= 1 0) (+ 1 (+ 2 2)) (f3 (ite (<= (- 1 1) 0) (+ 1 (+ 2 2)) (f3 (- (- 1 1) 1)))))) | (= (ite (<= (divf (+ 1 x) 2) 0) 
         j1 (v1 (- (divf (+ 1 x) 2) 1) (u2 (modf (+ 1 x) 2) 1) j1)) (v1 (divf x 2) (u2 (modf x 2) h2) v3))
47206642 (/\ (= (w1 (+ 1 x)) (u0 (+ 1 x) 1)) (= (- 2 (u0 x 1)) (- 2 (w1 x)))) | (= (v1 (divf x 2) (u2 (modf x 2) 1) v3) 
         (ite (<= 1 0) (+ 1 (+ 2 2)) (f3 (ite (<= (- 1 1) 0) (+ 1 (+ 2 2)) (f3 (- (- 1 1) 1)))))) | (= (ite (<= (divf (+ 1 x) 2) 0) 
         j1 (v1 (- (divf (+ 1 x) 2) 1) (u2 (modf (+ 1 x) 2) h2) j1)) (v1 (divf x 2) (u2 (modf x 2) h2) v3))
47234354 (/\ (= (w1 (+ 1 x)) (u0 (+ 1 x) 1)) (= (- 2 (u0 x 1)) (- 2 (w1 x)))) | (= (v1 (divf x 2) (u2 (modf x 2) 1) v3) 
         (ite (<= 1 0) (+ 1 (+ 2 2)) (f3 (ite (<= (- 1 1) 0) (+ 1 (+ 2 2)) (f3 (- (- 1 1) 1)))))) | (= (ite (<= (divf (+ 1 x) 2) 0) 
         j1 (v1 (- (divf (+ 1 x) 2) 1) (u2 (modf (+ 1 x) 2) 1) (u3 g3 h3))) (v1 (divf x 2) (u2 (modf x 2) 1) (u3 g3 h3)))
47741141 (/\ (= (u1 (h1 (+ 1 x)) (i1 (+ 1 x)) j1) (v0 (+ 1 x))) (= (+ z (v0 x)) (+ z (u1 (h1 x) (i1 x) j1)))) | (= (v1 (divf x 2) 
         (u2 (modf x 2) 1) v3) (ite (<= 1 0) (+ 1 (+ 2 2)) (f3 (ite (<= (- 1 1) 0) (+ 1 (+ 2 2)) (f3 (- (- 1 1) 1)))))) | (= (ite 
         (<= (divf (+ 1 x) 2) 0) j1 (v1 (- (divf (+ 1 x) 2) 1) (u2 (modf (+ 1 x) 2) 1) v3)) (v1 (divf x 2) (u2 (modf x 2) 1) v3))
47826876 (/\ (= (u1 (h1 (+ 1 x)) (i1 (+ 1 x)) j1) (v0 (+ 1 x))) (= (+ z (v0 x)) (+ z (u1 (h1 x) (i1 x) j1)))) | (= (v1 (divf x 2) 
         (u2 (modf x 2) 1) v3) (ite (<= 1 0) (+ 1 (+ 2 2)) (f3 (ite (<= (- 1 1) 0) (+ 1 (+ 2 2)) (f3 (- (- 1 1) 1)))))) | (= (ite 
         (<= (divf (+ 1 x) 2) 0) j1 (v1 (- (divf (+ 1 x) 2) 1) (u2 (modf (+ 1 x) 2) h2) j1)) (v1 (divf x 2) (u2 (modf x 2) h2) j1))
147491208 (= (s1 (+ 1 x)) (s1 x)) | (/\ (= (w1 (+ 1 x)) (v0 (+ 1 x))) (= (w1 x) (v0 x))) | (= j1 (s1 x)) | (<= x (v0 (v0 1)))
189948521 (= (s1 (+ 1 x)) (s1 x)) | (/\ (= (v0 (+ 1 x)) (w1 (+ 1 x))) (= (v0 x) (w1 x))) | (= j1 (s1 x)) | (<= x (v0 (v0 1)))
234621461 (= (s1 (+ 1 x)) (s1 x)) | (/\ (= (w1 (+ 1 x)) (v0 (+ 1 x))) (/\ (= (w1 x) (v0 x)) (= (w1 2) (s1 x)))) | (<= x (v0 (v0 1)))
259862681 (= (s1 (+ 1 x)) (s1 x)) | (/\ (= (w1 (+ 1 x)) (v0 (+ 1 x))) (/\ (= (w1 x) (v0 x)) (= (w1 2) (s1 x)))) | (<= x (v0 (w1 1)))
270905203 (= (s1 (+ 1 x)) (s1 x)) | (/\ (= (w1 (+ 1 x)) (v0 (+ 1 x))) (/\ (= (w1 x) (v0 x)) (= j1 (s1 x)))) | (<= x (v0 (v0 1)))
284249674 (= (s1 (+ 1 x)) (s1 x)) | (/\ (= (w1 (+ 1 x)) (v0 (+ 1 x))) (/\ (= (w1 x) (v0 x)) (= (s1 2) (s1 x)))) | (<= x (v0 (w1 1)))
365475391 (/\ (= (w1 (+ 1 x)) (v0 (+ 1 x))) (/\ (= (v0 x) (w1 x)) (/\ (= (s1 0) (s1 x)) (= (s1 (+ 1 x)) (s1 x)))))
775226841 (= (s1 2) (s1 x)) | (= (s1 (+ 1 x)) (s1 x)) | (/\ (= (v0 (+ 1 x)) (w1 (+ 1 x))) (= (v0 x) (w1 x)))
\end{verbatim}
\end{tiny}
\end{document}